\documentclass[letterpaper, 10 pt, conference]{ieeeconf}  

\IEEEoverridecommandlockouts                
\usepackage{amsmath}
\usepackage[ruled,vlined]{algorithm2e}
\usepackage{caption}
\usepackage{tabu}
\usepackage{booktabs}
\usepackage{etoolbox}
\usepackage{graphicx}
\usepackage{xcolor}
\usepackage{amsfonts}

\newcommand{\tocite}[1]{%
\textcolor{red}{[cite:\ifthenelse{\equal{#1}{}}{}{#1}]}
}

\usepackage[backend=biber,
            hyperref=true,
            url=false,
            isbn=false,
            doi=false,
            backref=false,
            style=ieee,
            natbib=true,
            mincitenames=1,
            maxcitenames=1,
            citestyle=numeric-comp,
            sorting=nyt,
            block=none]{biblatex}

\title{\LARGE \bf
Non-Markov Policies to Reduce\\ Sequential Failures in Robot Bin Picking
}

\author{Kate Sanders$^{1}$, Michael Danielczuk$^{1}$, Jeffrey Mahler$^{1}$, Ajay Tanwani$^{1}$, Ken Goldberg$^{1}$
\thanks{$^{1}$The AUTOLab at University of California, Berkeley. \{katesanders, mdanielczuk, jmahler, ajay.tanwani, goldberg\}@berkeley.edu}
}

\begin{document}

\maketitle
\begin{abstract}
A new generation of automated bin picking systems using deep learning is evolving to support increasing demand for e-commerce.  To accommodate a wide variety of products, many automated systems include multiple gripper types and/or tool changers.  However, for some objects, sequential grasp failures are common:  when a computed grasp fails to lift and remove the object, the bin is often left unchanged; as the sensor input is consistent, the system retries the same grasp over and over, resulting in a significant reduction in mean successful picks per hour (MPPH).  Based on an empirical study of sequential failures, we characterize a class of ``sequential failure objects" (SFOs) -- objects prone to sequential failures based on a novel taxonomy.  We then propose three non-Markov picking policies that incorporate memory of past failures to modify subsequent actions. Simulation experiments on SFO models and the EGAD dataset \cite{morrison2020egad} suggest that the non-Markov policies significantly outperform the Markov policy in terms of the sequential failure rate and MPPH. In physical experiments on 50 heaps of 12 SFOs the most effective Non-Markov policy increased MPPH over the Dex-Net Markov policy by 107\%.  
\end{abstract}

\section{Introduction} \label{sec:introduction}
State-of-the-art e-commerce bin picking systems are able to successfully grasp a wide variety of objects, but may fail repeatedly in a given situation, reducing mean picks per hour (MPPH)~\cite{ciocarlie2014towards}, \cite{lenz2015deep}, \cite{mahler2017learning}, \cite{zeng2018robotic}. For example, Dex-Net 4.0~\cite{mahler2019learning} has two grasping modalities (a vacuum suction cup gripper and a parallel jaw gripper) that offer 95\% reliability for standard objects, but can drop to 63\% reliability for objects labeled as ``adversarial" by \citet{mahler2019learning}. A smooth, porous object made of fabric may appear to be a good grasp candidate for the vacuum suction cup gripper (see Section~\ref{sec:failures}), but the fabric typically makes a secure vacuum seal infeasible. Since this property is not visible with a standard overhead depth camera, a bin picking system may repeatedly and unsuccessfully attempt to grasp the object using the vacuum suction cup gripper. We label objects that can result in repeated grasp failures for a given gripper type but can be grasped by at least one gripper type in the robot setup as \textit{sequential failure objects} (SFOs). Examples of bin picking failures are depicted in Figure 1.

\begin{figure}
\centering
\includegraphics[width=\linewidth]{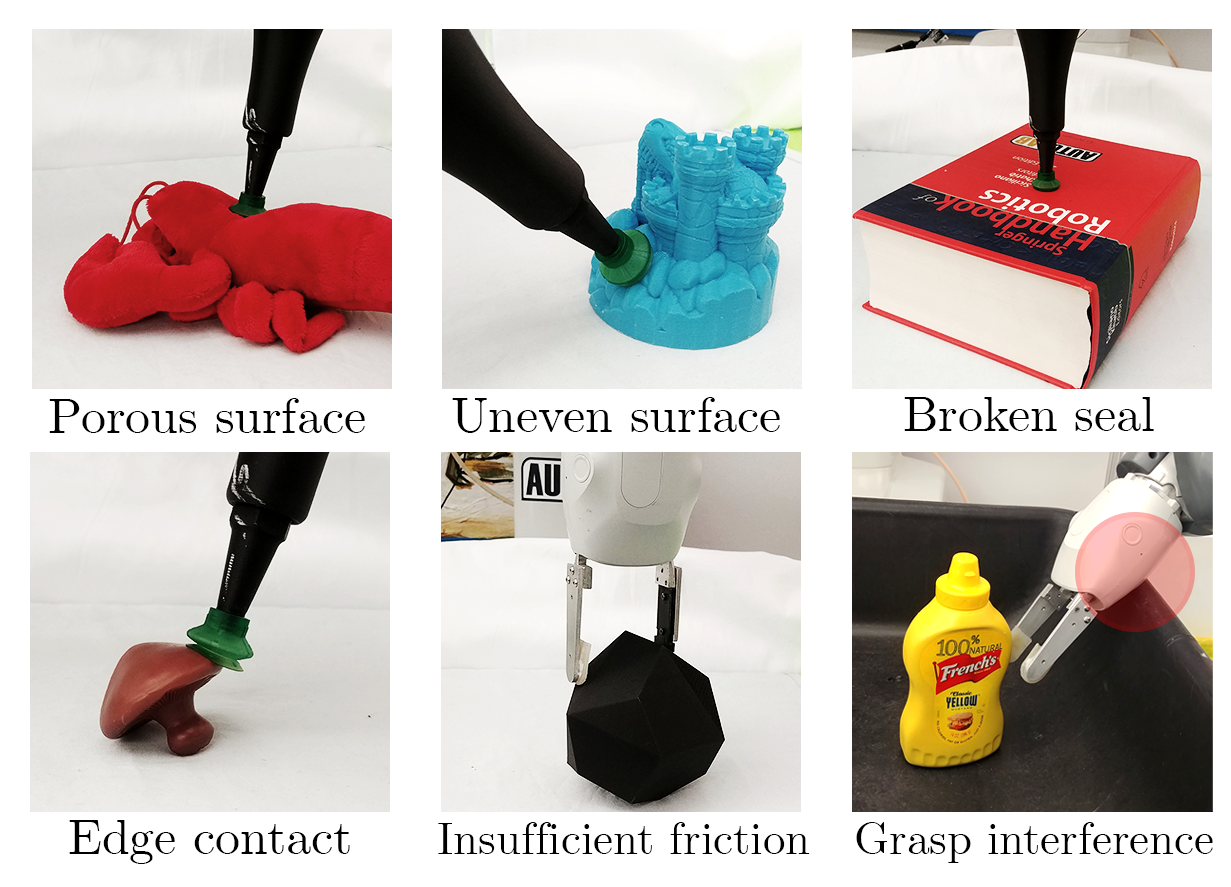}
\caption{Failure modes for a parallel jaw gripper and a vacuum suction cup. We define objects that result in repeated failures as sequential failure objects (SFOs) and aim to reduce the sequential failure rate (SFR) across these objects.}
\label{fig:failuremodes}
\end{figure}

When an object is not graspable by one gripper, one option is to grasp the object with a different gripper. Tool changers, commonly used in bin picking systems, can allow a robot to switch grippers to attempt different types of grasps on the object \cite{mahler2019learning}, \cite{gokler1997design}, \cite{kratky2019bin}, \cite{gyimothy2011experimental}. However, bin picking systems typically use Markov policies when grasping unknown objects, and choose an action using only the current observation (see Section~\ref{sec:relatedwork}). This lack of ``memory" can lead to sequential failures for SFOs, because unsuccessful actions may result in similar observations at the next state. In this paper, we explore how non-Markov policies that depend on the history of observations and actions can decrease the sequential failure rate (SFR) and increase mean picks per hour (MPPH) over a Markov policy.

In this paper, we first identify failure modes produced by SFOs by studying empirical data collected during physical bin picking experiments. We then introduce a set of three non-Markov policies to reduce failures caused by these SFOs. These policies are then tested in simulation to explore the applicability of the policies with respect to general bin picking systems. Finally, we evaluate the policies on a physical bin picking system.

 
This paper makes the following contributions:
\begin{enumerate}
\item A characterization of sequential failures in automated bin picking in terms of Sequential Failure Objects and Sequential Failure Rate.  
\item Three non-Markov picking policies that use past sensor data, actions, and rewards.
\item  Results from 8400 simulated  and 50 physical experiments suggesting that Non-Markov picking policies can provide significant increase in picks per hour for SFOs. 
\end{enumerate}

\section{Related Work} \label{sec:relatedwork}
\subsection{Bin Picking}
\citet{ellekilde2012applying} are one of the first to apply learning methods to improve grasping in bin-picking scenarios. \citet{ciocarlie2014towards}, \citet{lenz2015deep}, \citet{mahler2017learning}, and \citet{zeng2018robotic} use RGB or depth image observations at the current time step to generate grasps for bin picking via neural networks without consideration of past actions. \citet{zeng2018robotic} attempt to recognize training objects from a collection of potentially unseen objects, but if training objects cannot be recognized, they resort to using a Markov policy. \citet{levine2018learning} use a convolutional neural network to teach robots hand-eye coordination grasping policies.  \citet{quillen2018deep} explore deep reinforcement learning for bin picking, and emphasize the importance of generalization to novel objects in robotic bin picking. Aside from \citet{zeng2018robotic}, these methods all use Markov policies to compute grasps. In contrast, we consider non-Markovian bin picking processes.
Recent work includes research conducted by \citet{murali20196}, who explore picking single objects out of clutter by computing 6-DoF grasps, \citet{gabellieri2020grasp} who introduce a method for grasping unknown objects by leveraging human demonstrations, and \citet{pedersen2020grasping} who grasp unknown objects by training a deep neural network grasping agent on simulated data. \citet{bodnar2019quantile} introduce the Quantile QT-Opt deep reinforcement learning algorithm which can be used for risk-aware grasping.

We use Dex-Net 4.0 \cite{mahler2019learning}, a bin picking system that can grasp a wide variety of objects from a heap using a parallel jaw gripper and vacuum-based suction-cup gripper. It takes as input an overhead depth image of the heap and outputs a corresponding grasp action and grasp quality score. Dex-Net 4.0 achieves a high performance for a variety of unknown objects; on standard objects it achieves over 300 mean picks per hour and its reliability rate reaches above 95\%. However, on objects labeled by \citet{mahler2019learning} as ``adversarial", its reliability rate can drop to 63\%. As with the other systems mentioned, Dex-Net 4.0 uses a Markov bin picking policy.

\subsection{Uncertainty and Anomaly Detection in Automation}
Anomaly detection and classification are used to diagnose ways in which a robotic system can produce errors. \citet{srinivas1977error} introduces a ``failure tree" that can be built to analyze error and produce intelligent error recovery policies. \citet{chen2019novelty} detects novel states by using a salient network to learn an association between relevant input areas and predicted outputs.  \citet{park2017multimodal,park2018multimodal} use action-centric features such as torque and speed in addition to sensory input to classify encountered anomalies. \citet{wu2019latent} use a statistics-based multiclassifier to classify anomalies based on the current image observation. Bin picking failures, however, generally have similar sensory inputs regardless of underlying cause.

\subsection{Error Recovery in Automation}
Extensive work has also been done in the scope of error recovery. \citet{wu2018endowing} explore a policy by which the robot learns via demonstration after making an error. \citet{wang2019learning} use a multimodal transition model learned through reinforcement learning to improve the success rate of robots in unstructured environments. \citet{niekum2015learning} consider learning from human demonstrations that take advantage of hierarchical representations of the demonstrations. \citet{zhu2007reinforcement} and \citet{lee1983knowledge} use reinforcement learning methods to recover from errors that are caused by kinematic failures such as shut downs and collisions, and \citet{gordon2019learning} use online learning to correct robot failures in the context of robot-assisted feeding. \citet{kaipa2016addressing} address the handling of failure modes in robotic bin picking, but focus on failure modes associated with locating and placing objects and make use of human-aided perception and position planning.

Standard reinforcement learning approaches and learning through demonstration methods are not easy solutions to the problem we address in this paper. Reinforcement learning is difficult for picking novel adversarial objects because it is unclear if there is a correlation between depth images and SFO failure modes (see Section~\ref{sec:failures}). Similarly, learning through demonstrations is not generally feasible as it requires a teaching period for every error-prone object.

\section{Problem Statement} \label{sec:problemstatement}
We consider the problem of recovering from errors caused by SFOs (as opposed to grasp imprecision, hardware, etc.) given one or more grippers using a tool changer in the bin picking setting.


\subsection{Actions}
Let action $\alpha$ describe a tuple $(g,\theta,s)$ consisting of gripper $g\in \mathcal{G}$ where $\mathcal{G}$ is a discrete set of available grippers in a system, action parameter $\theta$, and RGBD image $s$. Let $r$ represent reward ($r=1$ if the grasp is successful, otherwise $r=0$). For instance, if $g$ is a suction or parallel jaw gripper, action parameter $\theta$ is a 3D position and orientation of the gripper defined by a tuple $(R,t)\in\text{SE}(3)$ consisting of rotation $R$ and translation $t$. Let $\tau$ represent a sequence of $N_\tau$ actions $[\alpha_0,\alpha_1...,\alpha_{N_\tau-1}]$ executed over time $T_\tau$ with cumulative reward $r_\tau$ for a bin containing $n_\tau$ objects. A trial $\tau$ ends when either (a) $r_\tau=n_\tau$, (b) no grasp actions can be computed, or (c) a sequence of twenty failed grasps in a row occurs.

\subsection{Objective} \label{subsec:objective}

The objective is to increase the mean picks per hour of a bin picking system by reducing the expected number of sequential failed picks. We define two variables to keep track of repeated failures. For a sequence $\tau$, $M_\tau$ represents the number of failed picks immediately followed by another failed pick, and $F_\tau$ represents the number of successful picks immediately preceded by a series of two or more failed picks.\\

We use three performance metrics, namely the sequential failure rate (mean number of sequential failures per picked object), median sequence length of sequential failures, and mean picks per hour (used by \citet{mahler2017learning}). These are defined as:
$$
\text{SFR}=\mathbb{E}\left[\frac{M_\tau}{r_\tau}\right] \;\;\; \text{MSL}=\mathbb{E}\left[{\frac{M_{\tau}+F_{\tau}}{F_{\tau}}}\right] \;\;\; \text{MPPH}=\mathbb{E}\left[\frac{r_{\tau}}{T_{\tau}}\right]
$$

$T_\tau$ is measured in hours. For simulated experiments, $T_\tau$ is calculated by summing the planning and execution times (in hours) for each pick across a trial. For physical experiments, $T_\tau$ is estimated as the number of picks multiplied by $T_{pick}$ hours, the estimated time per pick.
\section{Failure Taxonomy} \label{sec:failures}


\subsection{Failure Case Study}
We study 20 physical bin picking trials of 25 objects executed with the standard (Markov) Dex-Net 4.0 policy \cite{mahler2019learning}. Three object sets -- Level 1, Level 2, and Level 3 -- were used. The 20 total trials were divided into bins with Level 1 objects, bins with Level 1 and 2 objects, bins with Level 2 objects, and bins with Level 3 objects. Level 3 objects include SFOs. This was the experiment setup used by \citet{mahler2019learning}. A total of 76 failures occurred across 576 picks (86.8\% reliability).

\subsection{Empirical Failure Taxonomy}
The failures are recorded in Table~\ref{tab:dnfailures}. They are each classified as one of 6 error types across the two grippers using video footage of the trials.

Suction gripper errors occur when the gripper is unable to form a vacuum seal due to:
\begin{itemize}
    \item \textbf{Edge contact.} The gripper is at the edge or corner of the object.
    \item \textbf{Uneven surface.} The cup cannot form to the surface of the object at the grasping point.
    \item \textbf{Porous surface.} Unobservable holes exist in the target object, which make forming a vacuum seal impossible.
    \item \textbf{Broken seal.} The contact's torque arm is too large. This typically occurs when the object being grasped is too heavy for the vacuum seal to lift it.
\end{itemize}
Parallel jaw errors may occur due to:
\begin{itemize}
    \item \textbf{Insufficient friction.} The object slips out of the parallel jaws' grasp due to insufficient friction at the contacts.
\end{itemize}
Errors for either gripper may occur due to:
\begin{itemize}
    \item \textbf{Grasp interference.} The target object is unable to be grasped due to interference from nearby objects, usually partially on top of the target object.
\end{itemize}

These errors are depicted in Figure~\ref{fig:failuremodes}. We further classify the above error types into three gripper-agnostic failure modes:
\begin{enumerate}
    \item \textbf{Gripper Type Failure} Failures that occur due to the grasped object being ill-suited for the chosen gripper. This error category encompasses all porous objects and broken seals. It also includes uneven surfaces and insufficient friction errors (when these apply to the entire object). These can be corrected by using another gripper.
    \item \textbf{Gripper Placement Failure.} Failures that occur due to the placement of the grasp or the combination of grasp placement and gripper type. This category includes the latter group of errors that can fall into gripper type failures, when these properties do not apply to the entire object. Additionally, this error type encompasses all edge contact errors. These can be corrected by grasping at another location, or sometimes by using another gripper.
    \item \textbf{Gripper Collision Failure.}
    Failures that occur due to environmental factors. This category includes gripper interference with other objects. Solutions for these errors are not the main focus of this paper.
\end{enumerate}

\begin{table}[t!]
\vspace{5pt}
\centering
 \begin{tabu} to \linewidth {X[1.5]X[c]X[2c]}
 \multicolumn{3}{c}{\textbf{Gripper Errors}} \\
 \textbf{Gripper Type} & \textbf{\# of Failures} & \textbf{\% of Total Failures}\\
 \toprule
 Parallel Jaw & 25 & 33\%  \\
 Suction & 51 & 67\% \\
 \bottomrule
 \end{tabu}
 
 \vspace{10pt}
 
 \begin{tabu} to \linewidth {X[1.5]X[c]X[2c]}
  \multicolumn{3}{c}{\textbf{Suction-Specific Failure Modes}} \\
 \textbf{Failure Type} & \textbf{\# of Failures} & \textbf{\% of Suction Failures} \\
 \toprule
 Edge contact & 6 & 13\% \\
 Uneven surface & 22 & 43\% \\
 Porous surface & 10 & 20\% \\
 Broken seal & 12 & 23\% \\
 Grasp interference & 1 & 2\% \\
 \bottomrule
 \end{tabu}
 
 \vspace{10pt}
 
 \begin{tabu} to \linewidth {X[1.5]X[c]X[2c]}
 \multicolumn{3}{c}{\textbf{Parallel Jaw-Specific Failure Modes}} \\
 \textbf{Failure Type} & \textbf{\# of Failures} & \textbf{\% of Parallel Jaw Failures} \\
 \toprule
 Insufficient friction & 24 & 96\%  \\
 Grasp interference & 1 & 4\% \\
 \bottomrule
\end{tabu}
\caption{A breakdown of failure modes from 20 physical Dex-Net 4.0 trials with 25 objects each. The most common failures observed were uneven surfaces (when using a suction gripper) and insufficient friction (when using a parallel jaw gripper). These errors are depicted in Fig~\ref{fig:failuremodes}.}
\label{tab:dnfailures}
\end{table}

\subsection{Sequential Failure Objects}
Multiple descriptors have been used to define difficult-to-grasp objects in recent literature. \citet{mahler2019learning} use the terms ``adversarial" and ``pathological" based on empirical observations: ``adversarial" for objects that caused a high rate of failures, and ``pathological" for objects such as transparent objects that almost always fail due to perception. \citet{wang2019adversarial} use the term ``adversarial" analogously to adversarial images to describe objects with geometry that is actively designed to cause grasp failures almost always for one specific modality (i.e. parallel jaw grippers).

For clarity in this paper, we define a new class of objects specific to
a robot setup including perception sensor and available
gripper types (for example, a depth sensor with vacuum suction and parallel-jaw
gripper).  For a given robot setup, we define ``Sequential Failure Objects" (SFOs) as a set of objects that can be successfully grasped by one of the available gripper types but consistently produce errors in grasp type or grasp placement for another gripper type.  For example, as a depth sensor cannot distinguish between surfaces that are non-porous vs porous, objects with porous surfaces are SFOs for the setup example above.

For physical experiments in this paper, the robot setup consists of a depth sensor, a weight sensor, a vacuum suction gripper, and a parallel jaw gripper, as described in Section~\ref{sec:physexps}. Examples of SFOs for this setup are shown in Figure~\ref{fig:objectset}.
\section{Four Non-Markov Policies} \label{sec:policies}

\begin{figure}[t!]
\vspace{5pt}
\centering
\includegraphics[width=\linewidth]{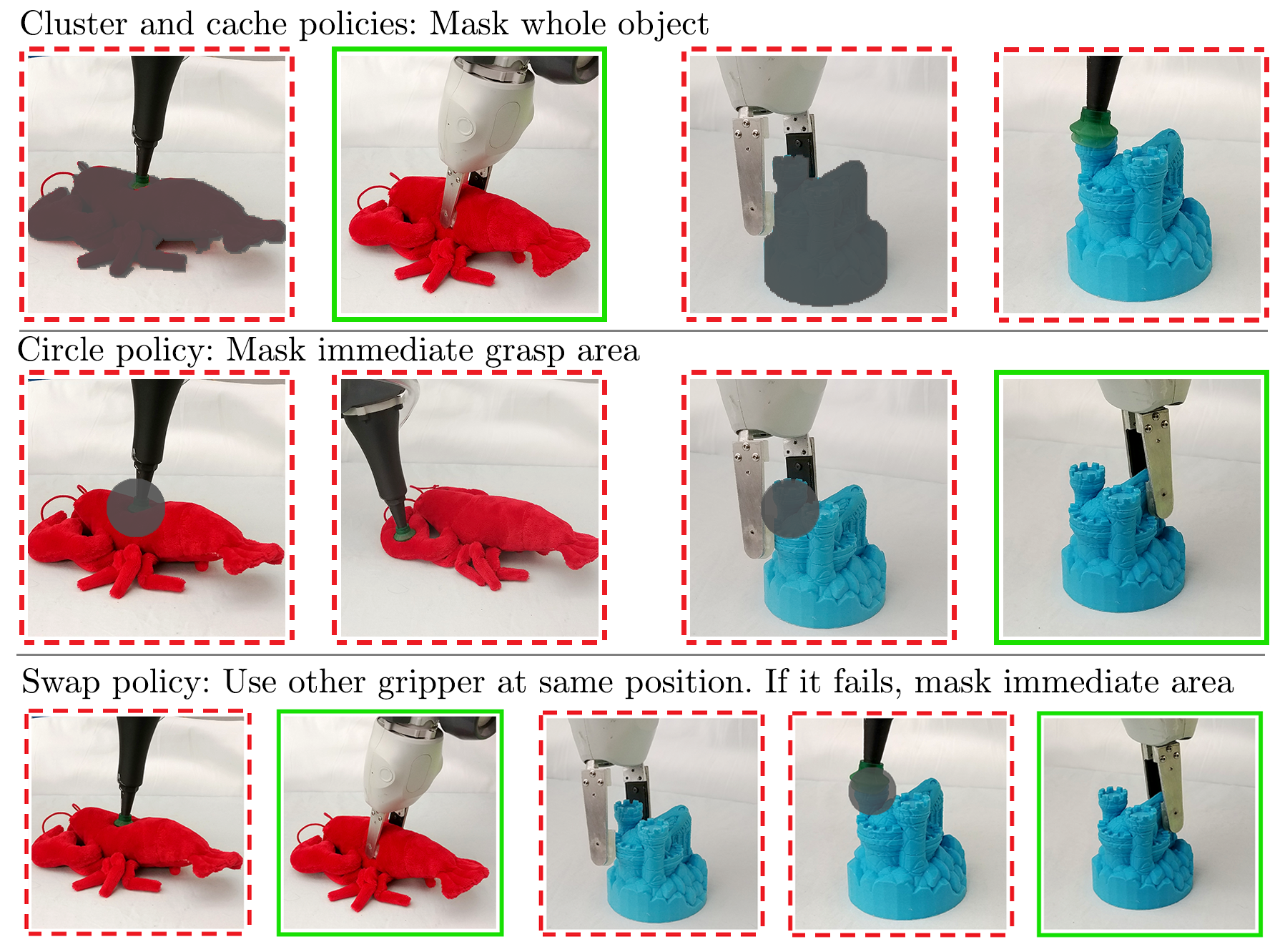}
\caption{Illustrations of the three non-Markov policies for a setup with two gripper types. Green solid borders around images denote successful grasps and red dashed borders denote unsuccessful grasps. The cache and cluster policies attempt to mask an entire object from a gripper's grasp space when it is unsuccessfully grasped, so a different gripper type must be used to grasp the object. In contrast, the circle policy only masks the immediate area, so the same gripper can be used on the object at another location. Examples of scenarios where these strategies succeed and fail are shown. The swap policy, shown in the bottom row, first attempts a similar grasp with another gripper. If the second grasp fails then the immediate area is masked in the same way it is masked by the circle policy.}
\label{fig:policies}
\end{figure}

We introduce three non-Markov policies to execute failure handling. They use masking to block off areas of specific grippers' grasp spaces in order to limit the range of grasps that can be executed, and use object tracking to determine what areas should be masked. An illustration of the three policies is shown in Figure~\ref{fig:policies}. All policies use actions generated by a grasp quality convolutional neural network (GQCNN)~\cite{mahler2019learning}, and all of the following policies begin by executing the best available grasps until an error occurs as determined by the weight sensor (detailed in Section~\ref{sec:physexps}).

\subsection{Cluster/Cache Policy}
The cluster \cite{mahler2019learning} and cache policies target gripper type failures (see Section~\ref{sec:failures}). When a gripper fails to grasp an object, that object is masked from that gripper's grasp space until it is picked by another gripper. To consistently mask the appropriate objects across time steps, the policy must segment the observation image into individual objects and track these objects across time steps.

The cluster policy uses the Euclidean clustering algorithm \cite{rusu2011point} to segment the observation image into objects. It tracks segmented objects across time steps by calculating a distance metric between pairs of current segmented objects and segmented objects from the previous time step. The pairs with the lowest distances are considered to be the same object and all metadata from the object at the previous time step will be propagated to its current state counterpart. The distance metric is calculated using features from a VGG-16 convolutional neural network~\cite{simonyan2014very}, segmentation size, and segmentation position.

The cache policy segments the current observation into objects using SD Mask R-CNN~\cite{danielczuk2019segmenting} and each segment is featurized using a ResNet35 CNN~\cite{he2015deep}. Objects that cause a grasp failure have their features (generated by the ResNet35 CNN) added to a failure cache. A distance metric is still used for tracking and is broken into two parts. First, a k-nearest neighbor algorithm matches featurized segments from the current state to a set of potential matches from the failure cache. Then, a Siamese network \cite{bertinetto2016fully} outputs a distance metric for each pair of potential matches. If the Siamese network score is below an $\epsilon$ threshold for a potential match, the current segment is masked from the grasp space for the gripper that had previously failed to grasp its paired object segmentation. The cache policy can preserve object metadata across bin picking trials because it does not use relative object position in its object tracking.

\subsection{Circle Policy}
The circle policy targets gripper placement errors, but can recover from gripper type errors as well. It assumes failures occur due to an incorrect grasping location, and restricts the chosen gripper from gripping at that location. It does not use object tracking. When a grasp fails, a circular mask of radius $r$ is placed at the center of the failed grasp location. The mask is removed if a different gripper successfully executes a grasp within the masked area at any point during the bin picking sequence. If subsequent grasps by that gripper on the object all fail, the circular masks will cover the entire object, and the policy must resort to using the other gripper -- the proposed strategy for recovering from gripper type errors.


\subsection{Swap Policy}
The swap policy also targets both error types. It first assumes the failure was a gripper type error by grasping with another gripper, and if the subsequent grasp fails, it resorts to the circle policy, which handles gripper placement errors.

After a failed grasp, the swap policy first attempts to grasp near the failure point with another gripper, and if this attempt fails, it then resorts to the circle policy, masking out both grippers' spaces with a circle of radius $r$. If the masks obfuscate the grasping spaces to the point where no grasps can be made, the radii of the masks are reduced by 50\% up to a threshold radius size $r'$.

\section{Simulated Experiments} \label{sec:simexperiments}

\subsection{Experiment Setup}
We evaluate the performance of the Markov, cluster, circle, and swap policies using two datasets of synthetic objects. The first consists of 1929 simulated SFOs, and the second consists of 100 adversarial objects from the EGAD dataset \cite{morrison2020egad}. The cache policy is not included because in the simulated environment we consider an optimal segmentation and tracking method, so it is functionally the same as the cluster policy. 

\subsubsection{SFO Dataset}
For each trial, we randomly select 12 objects (the number of objects used for physical experiments) from the 1929 synthetic objects for evaluation and artificially add random gripper type failures and gripper placement failures to these objects to simulate the properties of SFOs. The implementation of the failures are detailed below and the assignment of the failure properties are heap-independent. In our simulation, we use two identical suction grippers, but note that the properties given to the objects will cause each gripper to fail on different sets of objects and object locations, simulating two grippers of different types.

We use this dataset in three environments. They are differentiated by the types of failures that are simulated: only gripper type failures, only gripper placement failures, and a combination of both error types. These environments are detailed below:\\~\\
\textbf{Gripper Type Failures Only.} Each of the twelve objects are randomly chosen to be fully ungraspable by one of the available grippers (six objects per gripper in our setup). The radii associated with the circle and swap policies are both set at 0.015 meters, which is the same value used for radii in the physical environment.\\~\\
\textbf{Gripper Placement Failures Only.} Each of the twelve objects are only partially graspable by all available grippers. 30\% of adjacent faces of each object are randomly selected and coded to set the grasp reward to zero upon contact. These computations are deterministic for each object.\\~\\
\textbf{Both Failure Types.} Six objects are initialized as described in the gripper type failure environment (three objects per gripper in our setup) and six objects are initialized as described in the gripper placement failure environment.\\

\begin{figure}[t!]
\vspace{5pt}
\centering
\includegraphics[width=\linewidth]{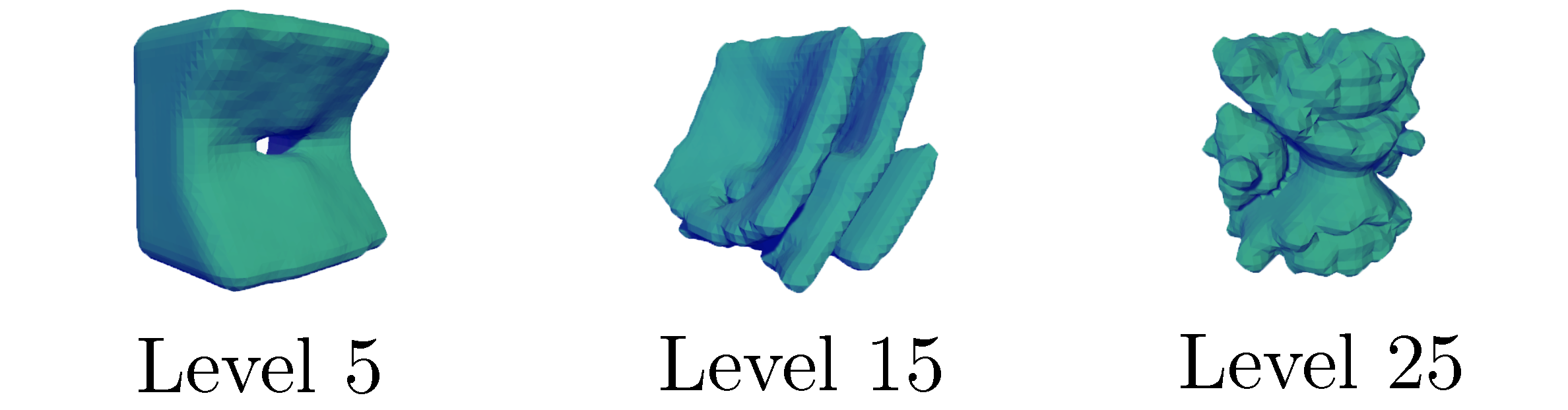}
\caption{Examples of objects from the EGAD dataset \cite{morrison2020egad} that were among the 100 meshes classified as the most difficult to grasp. These meshes were organized into 25 levels based on geometric complexity. The objects pictured above belong to geometric complexity levels 5, 15, and 25.}
\label{fig:egad}
\end{figure}

\subsubsection{EGAD Dataset}
The EGAD dataset, introduced by \citet{morrison2020egad}, is a collection of computer generated object meshes with the purpose of benchmarking grasping systems. The objects are divided into 25 ``grasping difficulty levels" based on \citet{wang2019adversarial}'s 75th percentile method. For this environment, 100 objects classified as the highest difficulty level are sampled to form heaps of 12 objects. These objects span evenly across 25 levels of geometric complexity based on metrics using angular deficits. The algorithm used to generate the objects encourages geometric diversity among objects at each complexity level. No modifications are made to these objects. For this environment, two different parallel jaw grippers are used.\\

We test the effectiveness of the policies detailed in the previous section over 500 trials each and for each environment. Policy parameters are not changed between environments. 

We prematurely end a bin picking trial if 20 errors occur in a row (as is done in the physical experiments due to time constraints) or if no more potential grasps can be found. In addition to the metrics described in Section~\ref{sec:problemstatement}, we also measure the percentage of objects in the heap that were successfully picked by the policy across all trials, the Percentage of Objects Successfully Picked (POSP).

\begin{table}[t]
\vspace{5pt}
\begin{tabu} to \linewidth {X[l]X[c]X[c]X[c]X[c]}
 \multicolumn{5}{c}{\textbf{SFO Objects: Gripper Type Failures Only}} \\
\textbf{Policy} & \textbf{SFR} & \textbf{MSL} & \textbf{MPPH} & \textbf{POSP}\\
 \toprule
 Markov & 1.71 & 3.0 & 789 & \textbf{0.69}  \\
 Cluster & 0.748 & \textbf{2.0} & \textbf{904} & 0.58 \\
 Circle & 0.753 & 3.0 & 893 & 0.68 \\
 Swap & \textbf{0.747} & \textbf{2.0} & 854 & \textbf{0.69} \\
 \bottomrule
 \end{tabu}
 
 \vspace{5pt}
 
 \begin{tabu} to \linewidth {X[l]X[c]X[c]X[c]X[c]}
  \multicolumn{5}{c}{\textbf{SFO Objects: Gripper Placement Failures Only}} \\
 \textbf{Policy} & \textbf{SFR} & \textbf{MSL} & \textbf{MPPH} & \textbf{POSP}\\
 \toprule
 Markov & 4.93 & 4.0 & 476 & 0.55 \\
 Cluster & \textbf{1.53} & \textbf{2.5} & \textbf{660} & 0.42 \\
 Circle & 2.05 & 3.0 & 659 & 0.58 \\
 Swap & 2.96 & 3.0 & 585 & \textbf{0.59} \\
 \bottomrule
 \end{tabu}
 
\vspace{5pt}
 
\begin{tabu} to \linewidth {X[l]X[c]X[c]X[c]X[c]}
  \multicolumn{5}{c}{\textbf{SFO Objects: Both Failure Types}} \\
 \textbf{Policy} & \textbf{SFR} & \textbf{MSL} & \textbf{MPPH} & \textbf{POSP}\\
 \toprule
 Markov   & 3.25 & 3.0 & 578 & 0.62 \\
 Cluster & \textbf{1.01} & \textbf{2.0} & \textbf{731} & 0.50 \\
 Circle & 1.22 & 3.0 & 700 & 0.61 \\
 Swap & 1.64 & 3.0 & 726 & \textbf{0.65} \\
 \bottomrule
 \end{tabu}
 
 \vspace{5pt}
 
\begin{tabu} to \linewidth {X[l]X[c]X[c]X[c]X[c]}
  \multicolumn{5}{c}{\textbf{EGAD Object Tests}} \\
 \textbf{Policy} & \textbf{SFR} & \textbf{MSL} & \textbf{MPPH} & \textbf{POSP}\\
 \toprule
 Markov   & 0.74 & 20.0 & 711 & 0.86 \\
 Cluster & \textbf{0.06} & \textbf{2.0} & \textbf{1137} & 0.88 \\
 Circle & 0.07 & \textbf{2.0} & 1087 & \textbf{0.92} \\
 Swap & 0.09 & \textbf{2.0} & 1077 & \textbf{0.92} \\
 \bottomrule
 \end{tabu}
 
\caption{Simulated experiment results across 500 trials per policy for each of the four environments described in Section VI using simulated SFOs and objects from the EGAD dataset \cite{morrison2020egad}. We compare the policies using four metrics: Sequential failure rate (SFR), median sequence length of sequential failures (MSL), mean picks per hour (MPPH) and percentage of objects successfully picked (POSP). All three non-Markov policies outperform the Markov policy in SFR and MPPH for each environment.}
\label{tab:simexps}
\end{table}

\subsection{Results}
Results are summarized in Table~\ref{tab:simexps}. Across all four environments, the non-Markov policies outperform the Markov policy in SFR and MPPH. While the cluster policy outperforms the circle and swap policies in the three performance metrics in the second and third environments, it achieves significantly fewer successful picks for the first three environments. The circle and swap policies achieve a similar number of successful picks or more than the Markov policy. We believe that the cluster policy achieves high metric values but a low number of successful picks because through masking out entire objects it eliminates successful grasps along with unsuccessful grasps. A further investigation into the relationship between mask size and number of successful picks is detailed in Section~\ref{sec:sensitivity}. This is not an issue in a physical environment where segmentation techniques are not as exact (and the objects are not fully masked by the policies). These simulation results indicate that both the circle and swap policies significantly improve SFR and MPPH regardless of object distribution while matching or increasing the number of successfully picked objects.

We believe that the results for the EGAD objects diverge from the other three environments because of the significant artificial difficulty introduced by the coded failure modes of the SFOs. Fewer failures were observed for the EGAD object experiments: The reliability ($\frac{\text{\# of successful picks}}{\text{\# of total picks}}$) of the three non-Markov policies was above 85\% for the EGAD dataset, while it was below 50\% for each of the policies in the other three environments. While the reliability scores of the EGAD dataset approximately align with the level 3 objects used by \citet{mahler2019learning}, the reliability scores of the SFO dataset are much lower.

\subsection{Circle Policy Radius Sensitivity} \label{sec:sensitivity}
To further explore the relationship between false positives and false negatives in grasp selection, we compare the efficacy of the circle policy as a function of its mask radius. For this experiment we use the Gripper Placement Failures Only environment and run 100 bin picking trials for each radius. Table~\ref{tab:circleablation} shows the results. These results support the hypothesis that there is a strong negative correlation between radius and SFR and between radius and POSP. This suggests that large masks eliminate many unsuccessful grasps, but also eliminate a number of successful grasps in the process.

\begin{table}[t]
\begin{tabu} to \linewidth {X[l]X[c]X[c]X[c]X[c]}
\multicolumn{5}{c}{\textbf{Circle Sensitivity Experiments}} \\
 \textbf{Radius} & \textbf{SFR} & \textbf{MSL} & \textbf{MPPH} & \textbf{POSP}\\
 \toprule
 0.005 & 2.48 & 3.0 & 471 & \textbf{0.55} \\
 0.015 & 1.79 & 3.0 & 475 & 0.53\\
 0.030 & 1.38 & \textbf{2.0} & \textbf{560} & 0.49\\
 0.045 & \textbf{1.12} & \textbf{2.0} & 548 & 0.44\\
 \bottomrule
\end{tabu}
\caption{Results illustrating the relationship between radius size and bin picking performance for the circle policy. Objects from the SFO dataset were used in the Gripper Placement Failures Only environment across 100 trials per radius size. There appears to be a negative correlation between radius and SFR and between radius and POSP. The radius is measured in meters.}
\label{tab:circleablation}
\end{table}

\section{Physical Experiments} \label{sec:physexps}
\begin{figure}
\centering
\includegraphics[width=\linewidth]{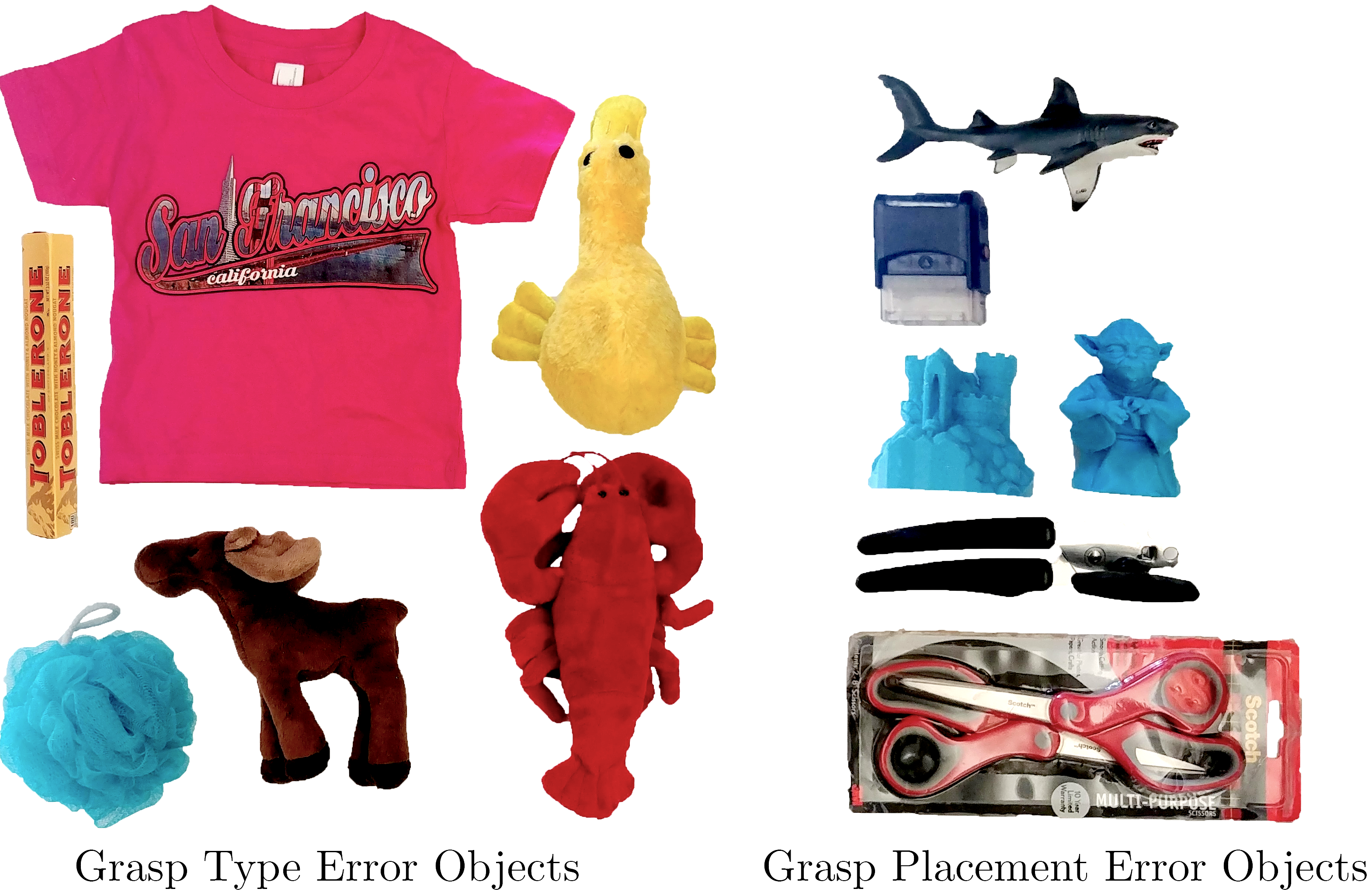}
\caption{12 SFOs used to benchmark the non-Markov bin picking policies compared to the standard Markov policy. SFOs that cause grasp type errors (left) are: a Toblerone chocolate bar, a shirt, a stuffed toy duck, a loofah, a stuffed toy moose, and a stuffed toy lobster. SFOs that cause grasp placement errors (right) are: a plastic shark, a rubber stamp, a 3D printed castle model, a 3D printed Yoda figure, a can opener, and a packaged set of scissors. These objects can be compared with the Level 1, Level 2, and Level 3 objects used to benchmark the development of Dex-Net 4.0 in \citet{mahler2019learning}.}
\label{fig:objectset}
\end{figure}

\begin{figure*}
\vspace{7pt}
\centering
\includegraphics[width=\linewidth]{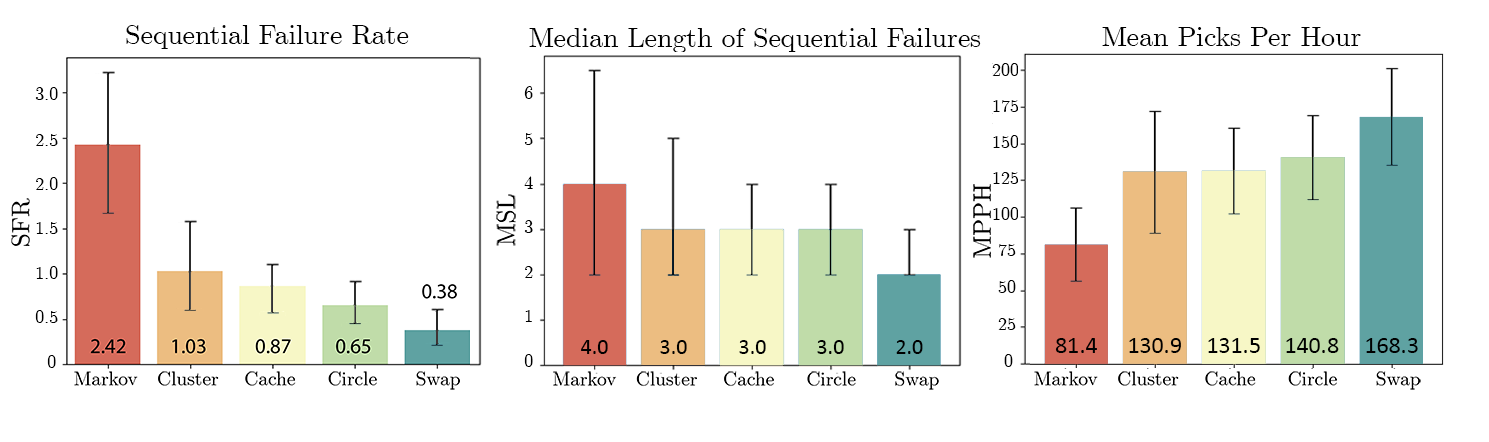}
\caption{Physical experiment results showing the (A) sequential failure rate, (B) median sequence length of sequential failures, and (C) mean picks per hour achieved by the five policies on the 12-object set detailed in section~\ref{sec:physexps} when run for ten trials each. The original Markov policy achieved the worst scores for each metric. The swap policy achieved the best scores for each metric, suggesting that it is the most successful policy of the five for this physical environment. The error bars show the standard error of the SFR and MPPH values and the upper and lower quartiles for the MSL.}
\label{fig:physresults}
\end{figure*}

\subsection{Experiment Setup}
We benchmark each policy (Markov, cluster, cache, circle, and swap) on a physical ABB YuMi bilateral industrial collaborative robot. The system is equipped with an overhead Photoneo PhoXi S industrial 3D scanner, a standard parallel-jaw gripper with silicone jaws, and a suction gripper with a 20mm diameter silicone suction cup. The bin containing the heaps of objects is mounted on a set of Loadstar load cells. Successful grasps are detected by the policy by measuring the change in weight detected by the load cells.

For each bin picking trial, the bin located below the ABB YuMi is filled with the set of 12 SFOs as shown in Figure~\ref{fig:objectset}. These objects are chosen due to the Markov policy's low average reliability (27\% compared to 95\% for standard objects~\cite{mahler2019learning}) when picking these objects from a bin. Each policy is then allowed to pick the objects iteratively from the bin, with grasping successes and failures recorded manually. If a policy attempted to execute a grasp that could not be completed due to a predicted joint collision (which occurred less than once per trial), the grasp was not recorded. Additionally, if an object is not successfully grasped by a policy after 20 failed grasp attempts, it was removed from the bin manually. This only occurred once across all trials, so we do not include the POSP as a metric for these experiments.\\

We measure policy effectiveness using the metrics introduced in Section~\ref{subsec:objective} across 10 trials per policy. One heap is defined as one trial. Results in Figure~\ref{fig:physresults} show that all four non-Markov policies outperform the Markov policy. We approximate time per pick as $T_{pick} = 12 \textrm{s} = \frac{1}{300}$ hours for the mean picks per hour metric, as each policy had similar execution times per pick (the setup used does not take additional time to switch tools).

\subsection{Sequential Failure Rate}
The swap policy reduced the SFR the most among the policies, which suggests that first attempting to use another gripper before masking out areas of a gripper's mask space shortens the length of the failure sequence more reliably than immediately masking grasping spaces. Despite access to previous image data and computationally intensive neural networks, the cache policy only performed marginally better than the cluster policy and worse than the simplest policies in terms of SFR and MPPH. We suspect that the uncertainty introduced by the networks used by the policy causes error to accumulate.

\subsection{Median Sequence Length of Sequential Failures}
The swap policy achieves the lowest median sequence length at 2 failures per sequence. Although the swap policy's MSL may be lower than the circle policy's MSL for Gripper Type Failures, it may be higher for Gripper Placement Failures due to it targeting Gripper Placement Failures first. Therefore, this metric may change depending on the distribution of SFOs used in the bin picking experiment.

\subsection{Mean Picks Per Hour}
The swap policy performs the best with regard to this metric. Conversely, the circle policy only achieves a MPPH that is 7.1\% higher than the cluster and cache policies despite achieving a much lower sequential failure rate and median failure sequence length, which suggests that while it reduces the length of failure sequences substantially, it does not reduce the total number of failures to such a degree. We hypothesize that the swap policy performed better than the circle policy due to the distribution of SFO types used in the experiment, as the swap policy frequently is able to grasp SFOs that cause gripper type failures with only one erroneous grasp, while the circle policy often executes multiple failed grasps on this object type.
\section{Conclusions and Future Work} \label{sec:futurework}
We explored three non-Markov policies for error recovery with an emphasis on mitigating failures caused by unobservable object properties.

We tested the effectiveness of the masking non-Markov policies on a set of 12 Sequential Failure Objects and compared their performances to that of the Dex-Net 4.0 Markov policy in simulation and on a physical robot. In simulation, the three non-Markov policies outperformed the Markov policy in terms of SFR and MPPH in three environments. Physical experiments suggest that the ``swap" policy can improve the sequential failure rate from 2.42 to 0.38, the median sequence length of sequential failures from 4.00 to 2.0, and the mean picks per hour from 81.4 to 168.3.

In future work, we will explore the potential for using a deep neural network to learn corrections to the predicted grasp qualities based on physical trials. We hypothesize that a reinforcement learning policy with access to many grasp attempts and their corresponding rewards on the same set of SFOs could learn to correct the predicted grasp qualities for these objects; however, learning these corrections at a large scale may be difficult, as there may not be a significant correlation between the failure modes a SFO may cause and its representation in a depth image.
\section*{Acknowledgements}
\footnotesize
This research was performed at the AUTOLAB at UC Berkeley in
affiliation with the Berkeley AI Research (BAIR) Lab. This research was supported in part by: NSF National Robotics Initiative Award 1734633 and by a Google Cloud Focused Research Award for the Mechanical Search Project jointly to UC Berkeley's AUTOLAB and
the Stanford Vision \& Learning Lab. The authors were supported in part by donations from Google and Toyota Research Institute, the National Science Foundation Graduate Research Fellowship Program under Grant No. 1752814, and by equipment grants from PhotoNeo and NVidia. We thank our colleagues who provided helpful feedback and suggestions, in particular David Wang, Vishal Satish, Matthew Matl, Jeff Ichnowski, Katherine Li, Ashwin Balakrishna, and Brijen Thananjeyan.

\renewcommand*{\bibfont}{\footnotesize}
\printbibliography
\end{document}